\documentclass[letterpaper]{article} 
\usepackage{aaai2027}  
\usepackage[hyphens]{url}  
\usepackage{graphicx} 
\usepackage{natbib}  
\usepackage{caption} 
\usepackage{algorithm}
\usepackage{algorithmic}

\usepackage{newfloat}
\usepackage{listings}
\DeclareCaptionStyle{ruled}{labelfont=normalfont,labelsep=colon,strut=off} 
\floatstyle{ruled}
\newfloat{listing}{tb}{lst}{}
\floatname{listing}{Listing}

\usepackage{booktabs}
\usepackage{amsmath}
\usepackage{amssymb}
\usepackage{cuted}

\title{GaussianSelector: Lightweight Human-Guided Object Selection\\in 3D Gaussian Splatting with Graph Optimization}
\author {
    Baihan Yang\textsuperscript{\rm 1}\equalcontrib,
    Tiexin Li\textsuperscript{\rm 2}\equalcontrib,
    Yuheng Liu\textsuperscript{\rm 3},
    Xin Lin\textsuperscript{\rm 1},
    Xinke Li\textsuperscript{\rm 2}\corresponding,
    Xiaohui Xie\textsuperscript{\rm 3},
    Truong Nguyen\textsuperscript{\rm 1}
}
\affiliations {
    \textsuperscript{\rm 1}UC San Diego\quad
    \textsuperscript{\rm 2}City University of Hong Kong\quad
    \textsuperscript{\rm 3}UC Irvine
}

\begin{document}

\maketitle

\begingroup
\renewcommand{\thefootnote}{}
\footnotetext{%
  \begin{tabular}{@{}r@{\hspace{0.4em}}l@{}}
    \textsuperscript{*} & Equal Contribution.\\
    \textsuperscript{\textdagger} & Corresponding Author.
  \end{tabular}%
}
\endgroup

\setlength{\stripsep}{-3pt}
\begin{strip}
  \centering
  \includegraphics[width=0.93\linewidth]{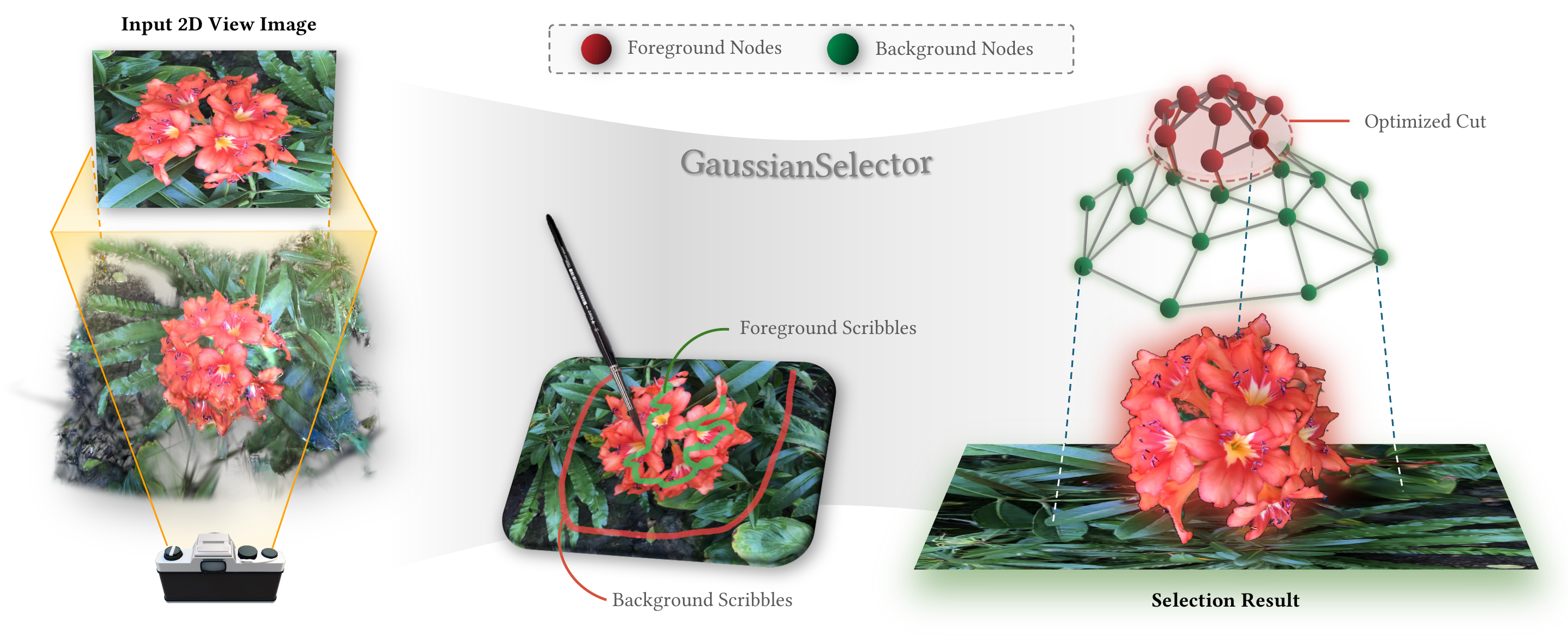}
  \captionof{figure}{
    Overview of \textbf{GaussianSelector}. We propose a plug-and-play interactive
    framework for object selection in 3DGS. Given user scribbles on a single
    view or a few views, GaussianSelector extracts 3D objects efficiently
    using graph-based optimization in the 3DGS space.
  }
  \label{fig:teaser}
  \vspace{1em}
\end{strip}

\begin{abstract}
Selecting a complete 3D object from a reconstructed scene with minimal user effort is essential for practical scene editing and embodied interaction. Existing 3DGS-based methods either retrain the Gaussian representation to embed per-object labels, or build dense multi-view SAM observations, both requiring heavy computation and dense viewpoint coverage that is rarely available in practice.
We present GaussianSelector, a training-free framework for interactive 3D object selection from sparse views and sparse scribble guidance. Operating directly on native Gaussian primitives, we coarsen dense Gaussians into geometrically coherent superpoints and construct a continuity-weighted graph using appearance and spatial cues. Sparse user scribbles are lifted into 3D via visibility-aware transmittance coverage, and selection is solved as a global graph-cut energy minimization that propagates sparse evidence to a complete 3D object. This design naturally supports multi-round refinement, where users iteratively correct the selection from additional viewpoints to progressively improve the result.
Experiments demonstrate that GaussianSelector achieves competitive selection quality against state-of-the-art multi-view SAM-based methods, while requiring significantly fewer interaction views and substantially lower computational overhead. These properties make it well suited for human-in-the-loop 3D scene editing and 3D asset extraction in real-world deployment scenarios.
\end{abstract}


\section{Introduction}

Interactive object selection in reconstructed 3D scenes is a fundamental capability for scene editing, asset extraction, robotic manipulation, and embodied perception \cite{kobayashi2022decomposing, miao2025towards, ren2022neural, yan20243dsceneeditor}.
Among recent 3D data formats, 3D Gaussian Splatting (3DGS) has become particularly attractive as it offers high-fidelity rendering together with an explicit set of 3D primitives \cite{kerbl20233d, lin2025hqgs, song2025d}.
This explicit structure makes 3DGS a promising substrate for post-hoc scene understanding: once a scene has been reconstructed, users should ideally be able to quickly select an object of interest and turn it into an editable 3D asset.

However, existing 3DGS-based interactive methods typically rely on user scribbles from multiple views and employ pretrained 2D foundation models, such as SAM \cite{kirillov2023segment}, to generate per-view segmentation masks.
These masks are then propagated back to the 3D Gaussian representation to enable interaction, often through feature indexing or related mechanisms, while some methods further require retraining the 3DGS backbone.
Such pipelines are therefore not only computationally expensive, but also heavily dependent on dense multi-view observations and user interaction across views.
In real-world scenarios, collecting such multi-view observations together with per-view scribbles is often impractical.
Moreover, occlusion, viewpoint changes, and appearance variations across views \cite{ren2022neural} can easily introduce inconsistencies among the generated prompts or masks, resulting in discontinuous and ambiguous 3D selection.
These limitations motivate us to study \textit{sparse scribble} interactive object selection, where only a single or a few spatially sparse yet informative views and scribbles are available \cite{ren2022neural}.

In this work, we present GaussianSelector, a lightweight Gaussian-native framework for interactive object selection in reconstructed 3DGS scenes. Instead of relying on SAM-generated multi-view masks or learned 3D feature fields, our method directly operates on the original 3D Gaussian primitives and propagates sparse user intent within a compact structural abstraction of the scene.
Specifically, we first oversegment dense Gaussians into geometrically coherent superpoints using Gaussian-native appearance and structural cues, and then convert sparse user scribbles into dense foreground/background evidence through visibility-aware scribble lifting and appearance contrast modeling. Object selection is finally formulated as an energy minimization problem on the superpoint graph and solved efficiently with graph cuts.
As a result, GaussianSelector is neural network-free and plug-and-play: it does not require pretrained segmentation networks during interaction, nor does it require retraining, finetuning, or refining the reconstructed 3DGS representation.
This design naturally supports iterative human-in-the-loop refinement, where users progressively correct the selection from additional viewpoints to improve the result.
This combination of sparse-view efficiency and iterative refinement makes our method particularly practical for real-world human-in-the-loop and embodied interaction scenarios.

Our main contributions are as follows:
\begin{itemize}
    \item We propose \textbf{GaussianSelector}, a lightweight and training-free framework for interactive object selection in reconstructed 3DGS scenes. Our method directly operates on native 3D Gaussian primitives, supports sparse scribble guidance from one or only a few interaction views, and does not require pretrained segmentation networks or any retraining, finetuning, or refinement of the underlying 3DGS representation.

    \item We introduce a Gaussian-native selection pipeline tailored to 3DGS scenes. Specifically, we design a geometrically coherent superpoint abstraction for dense Gaussian primitives, construct a structured graph over superpoints using Gaussian appearance and geometry cues, and develop a visibility-aware mechanism that lifts sparse user scribbles into superpoint-level foreground/background evidence for graph-based object selection.

    \item Extensive experiments on the NVOS benchmark show that GaussianSelector achieves competitive selection quality against state-of-the-art multi-view SAM-based methods, while requiring significantly fewer interaction views, lower VRAM consumption, and substantially less computation.
\end{itemize}
\section{Related Works}

\paragraph{3DGS Segmentation by Semantic Feature Field Learning.}
The explicit and differentiable nature of 3DGS enables segmentation via feature training directly in the 3D space using supervision from multi-view rendered images.
LangSplat~\cite{Langsplat},
VLGaussian~\cite{VLGaussian},
Feature3DGS~\cite{zhou2024feature}, 
and N2F2~\cite{bhalgat2024n2f2} leverage 2D foundation models, including CLIP~\cite{radford2021learning}, DINO~\cite{caron2021emerging}, and LSeg~\cite{LSeg}, to distill semantic cues into Gaussian feature fields for object segmentation. %
Further studies ~\cite{LEGaussian,qu2024goi,zuo2025fmgs,ji2025fastlgs} adopt discretization or compact embeddings to reduce storage while retaining semantic expressiveness.
Furthermore, SAGA~\cite{cen2025segment} and OmniSeg3D~\cite{ying2024omniseg3d} incorporate contrastive learning to better associate 3D objects with 2D SAM masks.
While these approaches demonstrate effectiveness in object retrieval and scene understanding in 3DGS scenes, they often introduce substantial computational overhead and costly labeling, which limit their real-world usability.

\paragraph{3DGS Segmentation by SAM Lifting.}
Driven by the advances in image segmentation, a line of work emerges that leverages the differentiable renderer of 3D Gaussian Splatting to directly lift 2D segmentation results into 3D space via cross-view consistency alignment to avoid the computational and annotation costs of large-scale feature field learning. Specifically, SA3D~\cite{SA3D}, SAGD~\cite{hu2024sagd}, GaussianGrouping~\cite{GaussianGrouping}, OpenGaussian~\cite{wu2024opengaussian}, FlashSplat~\cite{shen2024flashsplat}, VoteSplat~\cite{jiang2025votesplat}, and iSegMan~\cite{zhao2025isegman} directly transfer SAM masks into the Gaussian domain and enforce cross-view consistency to reduce discrepancies. 
Although these methods leverage strong 2D priors to achieve fine-grained segmentation and are more efficient than dense per-Gaussian feature training, they inherit ambiguity from cross-view lifted supervision and incur large VRAM memory cost in multi-view prediction.

\paragraph{3D Segmentation via Graphs.}
Given the effectiveness of graph structures in modeling spatial neighborhood relationships, many methods have explored leveraging graph-based formulations for object segmentation across different 3D representations.
NVOS \cite{ren2022neural} and GaussianCut \cite{jain2024gaussiancut} perform interactive object segmentation via graph-cut energy minimization on neural rendering representations. However, these methods formulate fine-grained segmentation directly on the global 3D representation, leading to high optimization complexity.
Other approaches exploit superpoint graph as a representation backbone to enable more expressive and computationally efficient scene understanding and segmentation, including Open3DIS \cite{nguyen2024open3dis} and \cite{hui2022learning} for point clouds, and AG$^2$aussian~\cite{wang2025ag2aussian} and InstanceGaussian~\cite{li2025instancegaussian} in 3D Gaussian Splatting.
Unlike our method, these approaches still learn semantic embeddings to define superpoints. We instead depart from semantic learning and construct superpoints directly from intrinsic 3DGS attributes and geometric structures, resulting in a lightweight, training-free interactive segmentation framework.

\section{Method}
\label{sec:method}

\begin{figure*}[t]
    \centering
    \includegraphics[width=\textwidth]{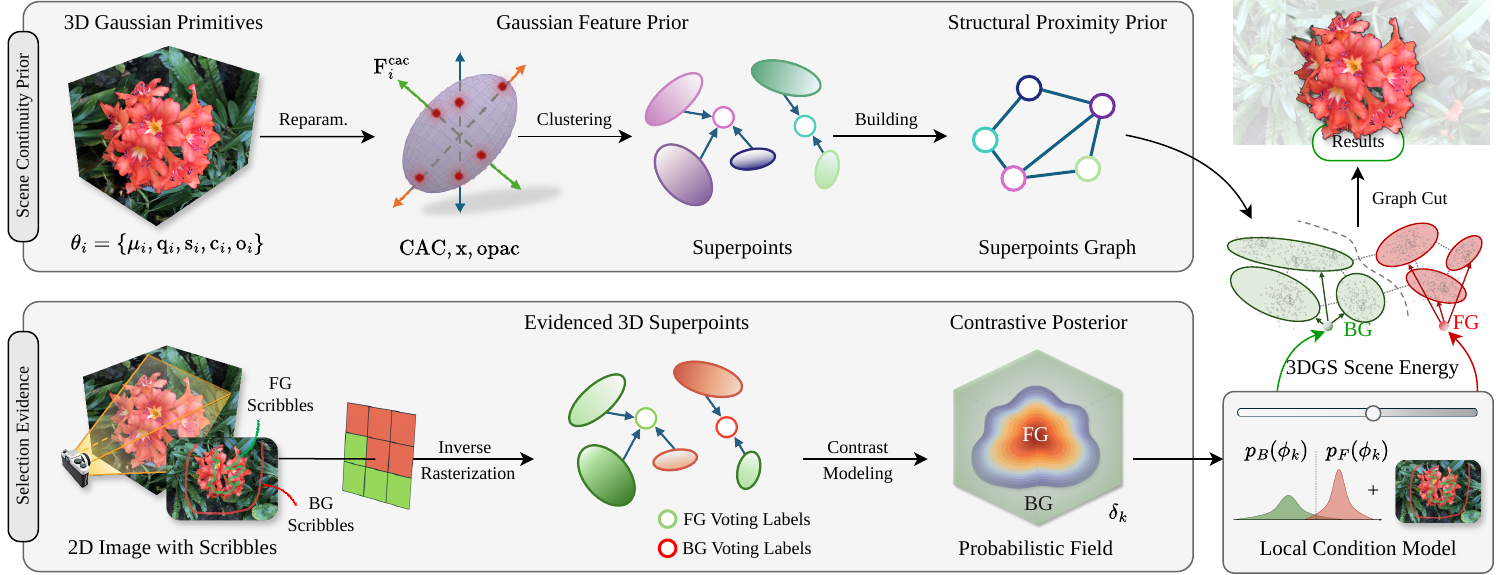}
    \caption{
Method overview. GaussianSelector first converts a reconstructed 3DGS scene into a superpoint graph using CAC appearance, opacity, and spatial proximity. Foreground (FG) and background (BG) scribbles are lifted to visible Gaussians via alpha-transmittance coverage and aggregated as sparse superpoint seeds. These seeds define contrastive foreground/background likelihoods, which are combined with the structural continuity prior in a binary graph-cut objective. The optimized superpoint labels are finally broadcasted back to Gaussian primitives to obtain the dense 3D selection.
}
    \label{fig:method}
\end{figure*}

Given a reconstructed 3DGS scene $\mathcal{G}=\{G_i\}_{i=1}^N$ and
sparse 2D scribbles $\mathcal{M}$, we seek a binary labeling
$L:\mathcal{V}_s\to\{F,B\}$ over scene-native superpoints that
maximizes the posterior
\begin{equation*}
  P(L \mid \mathcal{G}, \mathcal{M})
  \;\propto\;
  \underbrace{P(\mathcal{M} \mid L, \mathcal{G})}_{\text{how well }L\text{ explains the scribbles}}
  \;\cdot\;
  \underbrace{P(L \mid \mathcal{G})}_{\text{scene continuity prior}}.
\end{equation*}
Taking the negative log yields the energy we minimize:
\begin{equation}
  E(L)
  \;=\;
  \sum_{k\in\mathcal{V}_s} D_k(L_k)
  \;+\;
  \lambda
  \sum_{(i,j)\in\mathcal{E}_s} w_{ij}\,\mathbf{1}[L_i \neq L_j],
  \label{eq:energy}
\end{equation}
where the unary term $D_k$ encodes per-node foreground or background evidence
derived from the scribbles, and the pairwise term penalizes label
disagreement across edges weighted by scene continuity.
$E(L)$ is a binary submodular Potts energy with an exact global
minimum via s-t min-cut~\cite{boykov2006}.
The method instantiates this framework in three stages: scene
encoding (scribble-independent, cached once), broadcasting user intent
into seed evidence, and solving for the globally optimal labeling.
The overall algorithmic workflow is described in Algorithm \ref{alg:main}.

\subsection{Preliminary: Appearance Reparameterization for 3D Gaussian Splatting}
\label{sec:prelim}

Each primitive $G_i$ is parameterized by mean $\boldsymbol{\mu}_i\in\mathbb{R}^3$,
log-scale $\boldsymbol{s}_i\in\mathbb{R}^3$, rotation $R_i\in\mathrm{SO}(3)$,
opacity $\alpha_i\in(0,1)$, and spherical-harmonic (SH) radiance
coefficients $\boldsymbol{c}_i$.
Its spatial extent is given by the anisotropic covariance
\begin{equation}
  \Sigma_i \;=\; R_i\,\operatorname{diag}(\exp 2\boldsymbol{s}_i)\,R_i^\top.
  \label{eq:cov}
\end{equation}

Comparing Gaussians by appearance attributes is non-trivial: the DC
component of $\boldsymbol{c}_i$ ignores view-dependent effects, while raw SH
coefficients suffer from \emph{non-unique representation ambiguity}, the
same rendered appearance can be expressed by multiple distinct coefficient
vectors~\cite{xin2025learning}, making coefficient-space distances an
unreliable proxy for perceptual similarity. 
We characterize the appearance of each Gaussian as the \emph{Canonical Axis Color}
(CAC) feature, which is obtained by evaluating SH radiance along the six local canonical axes
$\mathcal{D}=\{\pm\mathbf{e}_x,\pm\mathbf{e}_y,\pm\mathbf{e}_z\}$
after transformation by the Gaussian's anisotropic support:
\begin{equation}
  \boldsymbol{f}_i
  \;=\;
  \operatorname*{concat}_{\boldsymbol{d}\in\mathcal{D}}\;
  \operatorname{SH}\!\left(
    \frac{R_i\operatorname{diag}(\exp\boldsymbol{s}_i)\,\boldsymbol{d}}
         {\|R_i\operatorname{diag}(\exp\boldsymbol{s}_i)\,\boldsymbol{d}\|_2+\epsilon}
    ;\;\boldsymbol{c}_i
  \right).
  \label{eq:cac}
\end{equation}
CAC can be understood as a \emph{moment projection} of the SH
radiance field onto the principal axes of the Gaussian's support: it
captures the first-order directional anisotropy of each primitive's
appearance while remaining deterministic and renderer-aligned.
The resulting 18-dimensional descriptor $\boldsymbol{f}_i$ serves as
the common currency for both graph structural abstraction
and appearance modeling throughout the method.

\subsection{Gaussian-Native Structural Proximity Prior}
\label{sec:graph}

Individual Gaussians may be under-constrained, partially transparent, or exhibit artifacts near object boundaries.
To obtain stable geometric boundaries for subsequent labeling, we represent 3D Gaussians as a spatial graph and perform graph coarsening to extract a compact structural abstraction.

We achieve this by aggregating them into \emph{superpoints} $\mathcal{S}=\{S_k\}_{k=1}^K$
via Leiden community detection on a  set of Gaussians, and
grouping primitives that share both spatial proximity and similar
appearance.
Each superpoint $S_k$ is represented by the mean-pooled position
$\bar{\boldsymbol{\mu}}_k$, CAC feature $\bar{\boldsymbol{f}}_k$,
and opacity $\bar{\alpha}_k$ of its members.
Crucially, superpoints serve as the low-frequency support of the
label field: the community structure groups Gaussians that should
share a label, so the subsequent graph-cut boundary falls naturally
at the seams between communities rather than across their interiors.

We build a $k$-NN graph $\mathcal{G}_s=(\mathcal{V}_s,\mathcal{E}_s)$
over superpoint centroids $\{\bar{\boldsymbol{\mu}}_k\}$.
Each edge $(i,j)$ carries a \emph{continuity weight}
\begin{equation}
  w_{ij}
  \;=\;
  \exp\!\left(-\frac{d_{ij}^2}{\sigma_{ij}^2}\right),
  \qquad
  d_{ij} = w_x\,d_x + w_c\,d_c + w_o\,d_o,
  \label{eq:affinity}
\end{equation}
where $d_x$, $d_c$, $d_o$ are normalized spatial, CAC, and opacity
distances, fixed hyperparameters $w_x+w_c+w_o=1$, and $\sigma_{ij}$ is \emph{self-tuned} from
the edge-distance distribution at local $k$-NN neighborhood $\mathcal{N}(i)$ to adapt to
Gaussian density:
\begin{equation}
\sigma_{ij} = \sqrt{\gamma_i \gamma_j}, \quad
\gamma_i = \mathrm{median}\{ d_{ik} \mid k \in \mathcal{N}(i)\},
\end{equation}

A gating mechanism removes edges whose $d_x$, $d_c$, or $d_o$ exceed the quantile threshold,
preventing outliers from distorting graph connectivity.
The continuity weight $w_{ij}$ directly instantiates the scene prior
$P(L\mid\mathcal{G})$: a label boundary crossing a high-$w_{ij}$ edge
incurs a large pairwise cost, concentrating cuts precisely where the
scene itself is discontinuous.
This graph constitutes a scribble-independent \emph{scene encoding}, computed once per session, and forms the fixed substrate over which
all subsequent inference runs.

\begin{algorithm}[t]
\caption{GaussianSelector}
\label{alg:main}
\begin{algorithmic}[1]
\STATE \textbf{Scene Encoding} \quad\textit{(once per scene)}
\STATE Compute CAC descriptors $\{\boldsymbol{f}_i\}$ for all Gaussians \hfill(Eq.~\ref{eq:cac})
\STATE Build $k$-NN Gaussian graph with weights $w_{ij}$ \hfill(Eq.~\ref{eq:affinity})
\STATE Cluster into superpoints $\mathcal{S}=\{S_k\}$ via Leiden
\STATE Build superpoint graph $\mathcal{G}_s=(\mathcal{V}_s,\mathcal{E}_s)$ over $\{\bar{\boldsymbol{\mu}}_k\}$
\STATE

\STATE Broadcast seeds to superpoints $\tilde{y}_k$; form $\mathcal{F},\mathcal{B}$ \hfill(Eq.~\ref{eq:seed})
\REPEAT
  \STATE Fit $p_F$ on $\mathcal{F}$,\; $p_B$ on $\mathcal{B}$;\; compute $\delta_k$ \hfill(Eq.~\ref{eq:contrast})
  \STATE Compute unary costs $D_k$ \hfill(Eq.~\ref{eq:unary})
  \STATE Minimize $E(L)$ via s-t min-cut;\; obtain $L^*$ \hfill(Eq.~\ref{eq:energy})
 \STATE Restrict to foreground subgraph: rebuild $\mathcal{G}_s$ on $\{k: L^*_k = F\}$;\; re-estimate $\mathcal{F}$, $\mathcal{B}$ on the restricted graph 
\UNTIL{$L^*$ converges or maximum iterations reached}
\STATE Broadcast $L^*$ to Gaussians;\; output $\hat{\mathcal{G}}$
\end{algorithmic}
\end{algorithm}
\subsection{Visibility-Aware Scribble Evidence Broadcast}
\label{sec:broadcast}

The user provides foreground and background scribbles $M^+$, $M^-$
on one or more rendered views.
Lifting these into 3D is subtle, as it operates over a soft volumetric visibility field rather than a hard spatial boundary: a Gaussian's projected center may
fall inside a scribble yet contribute negligibly to those pixels,
while a Gaussian lying outside may still bleed significantly through
its alpha-composited footprint.
We therefore define the \emph{visible coverage} of $G_i$ by scribble mask $M$
as the expected overlap under $G_i$'s own rendering distribution
$q_i(\boldsymbol{p})\propto\alpha_{i\boldsymbol{p}}T_{i\boldsymbol{p}}$:
\begin{equation}
  \rho_i
  \;=\;
  \mathbb{E}_{\boldsymbol{p}\sim q_i}\bigl[M(\boldsymbol{p})\bigr]
  \;=\;
  \frac{\sum_{\boldsymbol{p}}\alpha_{i\boldsymbol{p}}\,T_{i\boldsymbol{p}}\,M(\boldsymbol{p})}
       {\sum_{\boldsymbol{p}}\alpha_{i\boldsymbol{p}}\,T_{i\boldsymbol{p}}+\epsilon},
  \label{eq:coverage}
\end{equation}
where $T_{i\boldsymbol{p}}$ is the accumulated transmittance before
$G_i$.
Intuitively, $\rho_i$ asks: \emph{if one were to sample a pixel from
wherever $G_i$ actually renders, how likely is it to land inside the
scribble?}
Gaussians with foreground or background coverage above a per-view
threshold receive a seed label $y_i\in\{F,B\}$; ambiguous cases
remain unlabeled ($y_i=U$).
For multiple views, labels are reconciled by majority vote.

Seed labels are then broadcasted from Gaussians to superpoints by
majority aggregation:
\begin{equation}
  \tilde{y}_k
  \;=\;
  \operatorname{majority}\bigl\{y_i : i\in S_k\bigr\},
  \qquad \tilde{y}_k\in\{F,B,U\}.
  \label{eq:seed}
\end{equation}
This yields a sparse seed partition $\mathcal{F}=\{k:\tilde{y}_k=F\}$
and $\mathcal{B}=\{k:\tilde{y}_k=B\}$ that carries the user's intent
into the superpoint graph.

\subsection{Appearance Contrast Modeling} %
\label{sec:appearance}

With seeds $\mathcal{F}$ and $\mathcal{B}$ identified, we estimate
the likelihood term $P(\mathcal{M}\mid L,\mathcal{G})$ by learning
what each side of the selection looks like.
Each superpoint is described by the standardized feature
$\boldsymbol{\phi}_k=[z(\bar{\boldsymbol{f}}_k),\,z(\bar{\alpha}_k)]$,
combining CAC appearance and opacity; spatial information is
delegated to the pairwise term and excluded here.
We fit two independent Gaussian Mixture Models to the seed
features---$p_F$ on $\mathcal{F}$ and $p_B$ on $\mathcal{B}$---and
measure the \emph{radiance margin} of each superpoint as the
log-likelihood ratio between the two models:
\begin{equation}
  \delta_k
  \;=\;
  \log\frac{p_F(\boldsymbol{\phi}_k)}{p_B(\boldsymbol{\phi}_k)}.
  \label{eq:contrast}
\end{equation}
According to the Neyman--Pearson lemma, $\delta_k$ is the most powerful
test statistic for discriminating the foreground hypothesis from the
background hypothesis given $\boldsymbol{\phi}_k$: no other function
of the node's features carries more discriminative information about
its class.
A positive $\delta_k$ indicates that the node's appearance is better
explained by the foreground model; negative values point toward
background.
To obtain a calibrated and symmetric evidence field, we transform the raw log-likelihood ratio by a contrastive affine normalization:
\begin{equation}
  \hat{\delta}_k
  \;=\;
  s(\delta_k - m),
  \label{eq:contrastive_calibration}
\end{equation}
where $m$ denotes the midpoint between foreground/background seed medians in likelihood-ratio space, and $s$ is an adaptive scale determined by their separation.
Unlike raw per-class likelihoods, whose absolute scale depends on seed count
and covariance, $\delta_k$ is automatically centered by the
separation between $p_F$ and $p_B$.

The unary cost is derived from the calibrated contrastive posterior under equal class priors:
\begin{equation}
\begin{aligned}
D_k(c)
&=
-\log\frac{\exp(\operatorname{sg}(\delta_k,c))}
          {\exp(\delta_k)+\exp(-\delta_k)} \\
&\quad
+\beta\,\mathbf{1}[\tilde{y}_k\neq c,\; k\in\mathcal{F}\cup\mathcal{B}],
\end{aligned}
\label{eq:unary}
\end{equation}
where $\operatorname{sg}(\delta_k,c)=\delta_k$ if $c=F$ and
$-\delta_k$ if $c=B$.
The first term is the Bayes-optimal unary under equal priors; the
second reinforces seed superpoints toward their designated class with
weight $\beta$, encoding the user's direct evidence without treating
noisy lifted scribbles as hard constraints.
Substituting Eqs.~\eqref{eq:contrast}--\eqref{eq:unary} into
Eq.~\eqref{eq:energy} completes the instantiation of the MAP
objective.

\subsection{Graph-Cut Optimization and Interactive Inference}
\label{sec:optimization}

The energy $E(L)$ in Eq.~\eqref{eq:energy} is minimized exactly by
an s-t min-cut algorithm, which finds the globally optimal binary
partition of the superpoint graph.
The unary term pulls each node toward the class favored by its
appearance contrast; the pairwise term propagates this evidence
through the graph, aligning boundaries with the scene's own
continuity structure and suppressing isolated mislabelings caused
by noisy or incomplete seed coverage.
Once the optimal labeling $L^*$ is obtained, it is broadcasted back
to every Gaussian primitive: each $G_i\in S_k$ inherits $L^*_k$,
yielding the final dense foreground selection
$\hat{\mathcal{G}}=\{G_i:L^*_{k(i)}=F\}$.

GaussianSelector separates the scene encoding
(Secs.~\ref{sec:prelim}--\ref{sec:graph}) from the
scribble-dependent object modeling
(Secs.~\ref{sec:broadcast}--\ref{sec:optimization}).
This design makes our method well suited for efficient iterative object selection under interactive user inputs, as the large-scale Gaussian primitive graph is computed only once after scene loading.
When the user adds or corrects a scribble, only the evidence
broadcast, appearance modeling, and min-cut on the superpoint graph are rerun.
Our method therefore enables a natural human-in-the-loop workflow: an initial
sparse scribble yields a first selection; the user inspects the selection,
inputs a complementary scribble from an informative view, and the
system updates the dense 3D selection without modifying the scene
representation.
Since object selection is inherently local, we implement an ROI variant that restricts graph construction, optimization, and refinement to a local bounding box initialized from the scribble, enabling more efficient selection of small objects in large scenes.
\section{Experiments}
\label{sec:experiments}

\subsection{Experimental Setup}
\label{sec:exp_setup}

\paragraph{Dataset and Benchmark.}
We evaluate our method on the 8-task LLFF-NVOS~\cite{ren2022neural} benchmark and 5 scenes from 3D-OVS~\cite{3d-ovs} benchmark.
We further design two input settings for quantitative comparison on top of the initial NVOS scribble, including 1 additional round and 2 additional rounds of fine-grained user interactive refinement. %
This setup contextualizes our human-in-the-loop results in comparison to reported dense multi-view SAM-lifting-based results, while also highlighting the benefits of interactive refinement.
To provide a more comprehensive qualitative study across different 3DGS reconstruction settings, we further conduct qualitative experiments to visually demonstrate the object selection results.

\paragraph{Baselines and Metric.}
We compare our method with representative 3DGS segmentation methods including NVOS~\cite{ren2022neural}, FlashSplat~\cite{shen2024flashsplat}, GaussianCut~\cite{jain2024gaussiancut}, GaussianGrouping \cite{GaussianGrouping}, SAGA \cite{cen2025segment}, iSegMan \cite{zhao2025isegman}, OmniSeg3D \cite{ying2024omniseg3d}.
Following prior methods, we report intersection-over-union (IoU) between the rendered evaluation-view masks induced by selected Gaussians and the ground-truth object masks. We further compare the runtime of our method with the baselines to evaluate computational efficiency.

\paragraph{Implementation Details.}
All experiments are conducted on a workstation with an AMD Ryzen 9 9950X3D CPU and an NVIDIA V100 GPU.
We set parameters in a default setting, with $3$ GMM components for appearance contrast modeling, and $k=8$ for $k$-NN graphs.
We use a $0.95$ quantile for edge gating, a target seed confidence of $0.95$ (Eq.~\ref{eq:contrastive_calibration}), and a seed-evidence weight $\beta=4.0$ (Eq.~\ref{eq:unary}).
GaussianSelector has three user-adjustable hyperparameters tailored to different object selection tasks: the over-segmentation resolution which controls the granularity of the superpoint abstraction, and a connected component filtering threshold applied after graph-cut optimization to remove isolated regions, and a scale-outlier criterion that rejects excessively large Gaussian primitives, which are typically caused by highlight or shadow artifacts. The immediate visual feedback enables users to efficiently tune these settings to enhance the selection quality.

Our method supports multi-round interactive object selection across different viewpoints between the user and the model. The CAC features and superpoint graph construction are computed only once. During each subsequent interaction round, only the visibility-aware scribble lifting, seed-conditioned likelihood estimation, and graph-cut inference steps are re-executed, enabling efficient iterative refinement.
Most computations of our method are performed on the CPU.
More details are provided in the supplementary.

\subsection{Quantitative Results}
\label{sec:main_results}

\begin{table}[t]
\small
\centering
\setlength{\tabcolsep}{1mm}
\begin{tabular}{l|c|c|c|c|c}
\toprule
\textbf{Method} & \textbf{View} & \textbf{SAM} & \textbf{GPU} & \textbf{mIoU}$\uparrow$ & \textbf{Time}$\downarrow$ \\
\midrule

\multicolumn{4}{l}{\emph{Single-Round Settings}} \\

NVOS
& 1
& $\times$
& $\checkmark$
& \underline{70.1}
& -- \\

Ours w/ NVOS Scribble
& 1
& $\times$
& $\times$
& \textbf{85.3} 
& \textbf{0.2} \\

\midrule

\multicolumn{4}{l}{\emph{Multi-Round Settings}} \\

FlashSplat
& ALL
& $\checkmark$
& $\checkmark$
& 91.8 
& 0.8 \\

GaussianCut
& ALL
& $\checkmark$
& $\checkmark$
& \textbf{92.5} 
& 2.1 \\

GaussianGrouping
& ALL
& $\checkmark$
& $\checkmark$
& 90.6 
& 27.8 \\

SAGA
& ALL
& $\checkmark$
& $\checkmark$
& 90.9 
& 27.4 \\

iSegMan
& ALL
& $\checkmark$
& $\checkmark$
& 92.0 
& \underline{0.6} \\

OmniSeg3D
& ALL
& $\checkmark$
& $\checkmark$
& 91.7 
& 51.1 \\

Ours w/ 2 Rounds
& 2
& $\times$
& $\times$
& 89.6 
& \textbf{0.2} \\

Ours w/ 3 Rounds
& 3
& $\times$
& $\times$
& \underline{92.2} 
& \textbf{0.2} \\

\bottomrule
\end{tabular}
\caption{
Quantitative comparison on the NVOS benchmark.
We categorize methods by interaction setting, and explicit reliance on SAM supervision and GPU.
2/3 Rounds denotes 1/2 additional rounds of interactive refinement based on the scribble setting.
Our implementation does not require GPU beyond the 3DGS optimization/rendering.
The best results are highlighted in bold, and the second-best results are underlined.
Runtime is reported in \textbf{minutes}. "$-$" indicates unavailable.
}
\label{tab:main_results}
\end{table}

\begin{table}[t]
\centering
\small
\setlength{\tabcolsep}{1.5mm}
\begin{tabular}{l|ccccc|c}
\toprule
\textbf{Method} & \textbf{bed} & \textbf{bench} & \textbf{room} & \textbf{sofa} & \textbf{lawn} & \textbf{mIoU}$\uparrow$ \\
\midrule
SAGA & \textbf{97.4} & 88.4 & 81.0 & 71.2 & 94.8 & 86.5 \\
GSGrouping & \underline{97.3} & 73.7 & 79.0 & 68.1 & \textbf{96.5} & 82.9 \\
GaussianCut & 96.8 & \textbf{95.4} & \textbf{94.9} & \textbf{95.0} & 89.8 & \textbf{94.4} \\
Ours & 94.2 & 94.5 & \underline{92.0} & 94.0 & \underline{91.4} & 93.2 \\
Ours-ROI & 92.5 & \underline{94.8} & 91.9 & \underline{94.3} & 92.6 & \underline{93.6} \\
\bottomrule
\end{tabular}
\caption{
Quantitative comparison of interactive object selection methods on the 3D-OVS benchmark.
The best results are in bold, and the second-best results are underlined.
}
\label{tab:3dovs_results}
\end{table}

Table~\ref{tab:main_results} compares GaussianSelector with prior 3DGS object selection methods under different view and supervision settings, including single-view versus multi-view input and whether SAM mask supervision is required.

\paragraph{Single-Round Setting.}
We begin with the simplest and most computationally constrained interaction setting, where only a single view with NVOS scribbles is available for object selection. The representative 3D segmentation baseline under this setting is NVOS \cite{ren2022neural}.
With only the spatially sparse scribbles provided by the NVOS benchmark as input, GaussianSelector achieves 85.3 mIoU, significantly outperforming the NVOS baseline, which obtains 70.1 mIoU.
This result indicates that, even without multi-view mask propagation or pretrained 2D segmentation models, a Gaussian-native formulation can already recover high-quality 3D object selections from minimal user input.

\paragraph{Multi-Round Refinement Setting.}

We further evaluate GaussianSelector with extra rounds of scribble human-in-the-loop refinement at different viewpoints.
With one additional refinement view, our method improves to 89.6 mIoU, and with 2 additional rounds of interactive refinement on novel views, it further reaches 92.2 mIoU, becoming comparable to strong baselines built on dense-view priors.
While these baselines rely on SAM-derived cues across densely sampled views, whereas GaussianSelector achieves competitive performance with substantially fewer but informative views, selected online by users (1 per round), and without explicit requirement of image segmentation models.
As we allow users to select informative views and inject scribbles, our method mitigates the cross-view semantic ambiguity that often arises in dense multi-view lifting approaches.
The result also validates our design: directly exploiting the structural and appearance cues encoded in 3D Gaussian primitives is sufficient to support strong interactive selection.

\subsection{Efficiency Analysis}%
\label{sec:runtime}

We report the average runtime of each method over NVOS tasks in Table~\ref{tab:runtime_comparison} as a measure of computational efficiency.
A key advantage of GaussianSelector is that it decouples scribble-dependent interaction from scene-level optimization. This design enables efficient multi-round interactive scribble refinement and graph-cut optimization with low runtime overhead, making our method well suited for human-in-the-loop object selection.
In contrast to prior methods that require expensive training or high inference costs, GaussianSelector introduces much smaller inference overhead during subsequent human refinement. This observation demonstrates that our method enables high-quality human-in-the-loop object selection with minimal computational overhead.
Furthermore, since our method operates with significantly fewer input views than prior approaches and does not rely on deep neural networks such as SAM, it substantially reduces VRAM memory consumption and overall computational overhead.

\begin{figure}[t]
    \centering
    \includegraphics[width=1\linewidth]{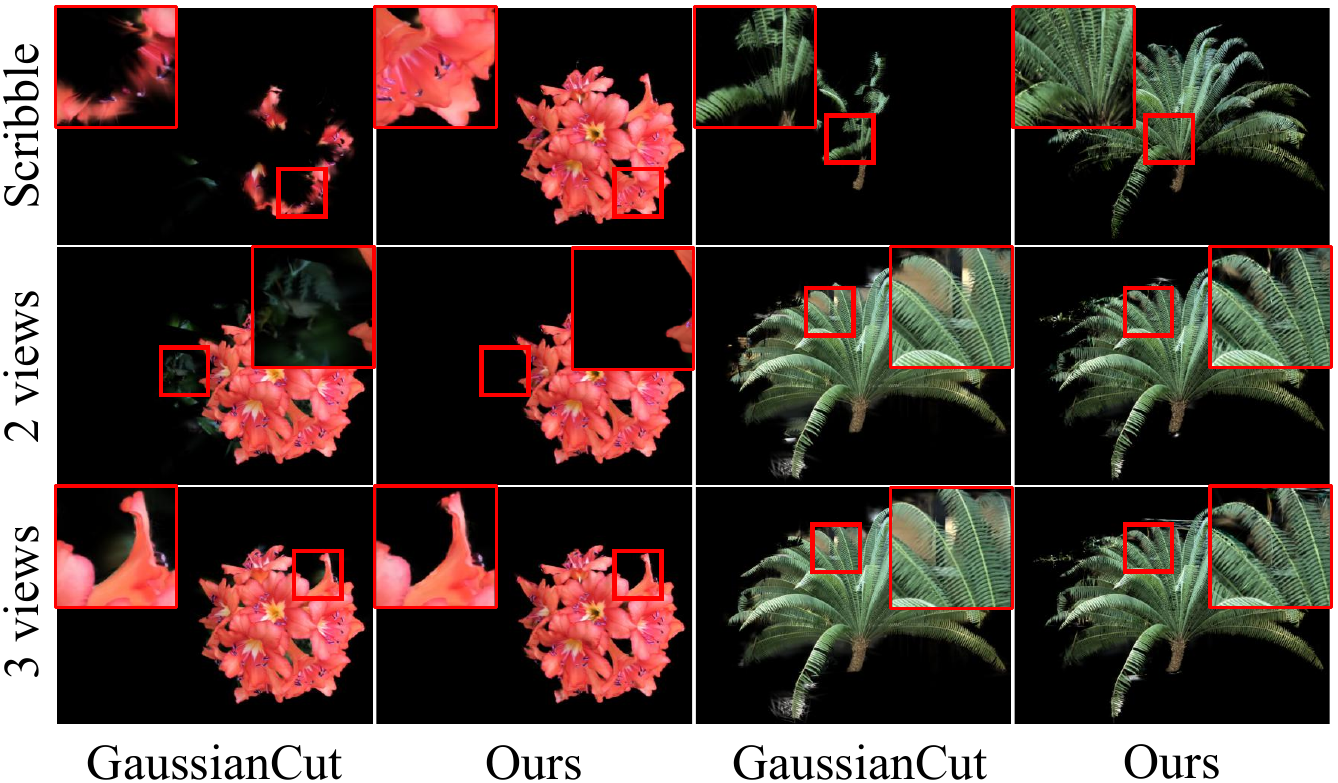}
    \caption{
    Qualitative comparison on NVOS under various input view settings (NVOS scribbles, 1/2 additional views or interaction rounds).
    }
    \label{fig:comp2}
\end{figure}

\begin{figure}[b]
    \centering
    \includegraphics[width=1\linewidth]{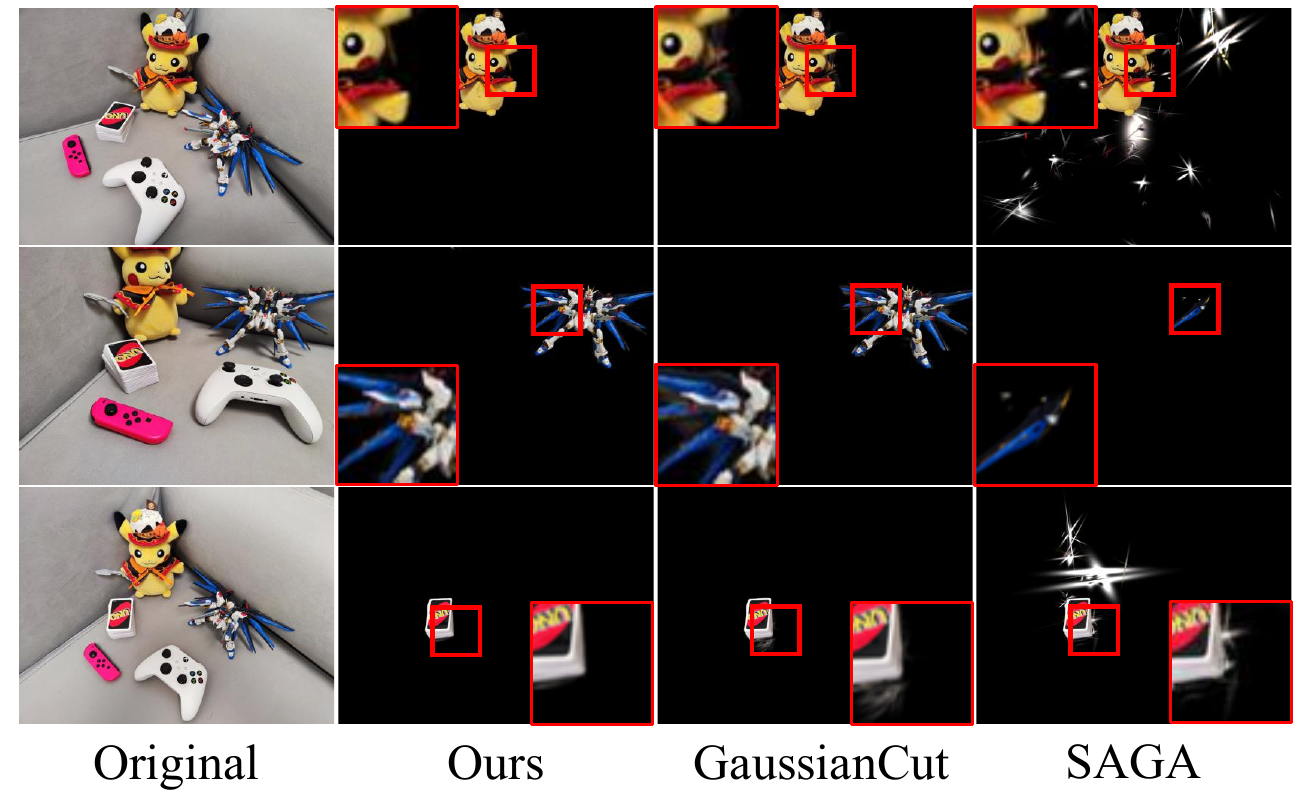}
    \caption{
    Qualitative comparison on 3D-OVS. Our method uses 5 interaction rounds; all baselines use all views.
    }
    \label{fig:comp3}
\end{figure}

\begin{table}[h]
\centering
\small
\begin{tabular}{l|c|c|c|c}
\toprule
\textbf{Method} & \textbf{Prep.} & \textbf{Training} & \textbf{Infer.} & \textbf{Total}$\downarrow$ \\
\midrule
FlashSplat      & 36.3  & --      & 13.1  & 49.4 \\
GaussianGrouping    & 159.7 & 1491.0  & 16.4  & 1667.1 \\
SAGA            & 190.2 & 1407.3  & 46.9  & 1644.4 \\
OmniSeg3D       & 89.6  & 2960.2  & 16.7  & 3066.5 \\
GaussianCut     & 35.7  & --      & 90.8  & 126.5 \\
\midrule
Ours & 11.3   & --      & 0.3  & \textbf{11.6} \\
Ours-ROI & 3.0 & -- & 0.2 & \textbf{3.2} \\
\bottomrule
\end{tabular}
\caption{
Average runtime comparison on all NVOS object selection tasks (in seconds).
Preparation denotes scene-level preprocessing or optimization.
Inference refers to the query stage.
Training denotes feature field learning when required.
}
\label{tab:runtime_comparison}
\end{table}

\subsection{Ablation Studies}
\label{sec:component_ablation}

\begin{table}[h]
\small
\centering
\begin{tabular}{c|c|c|c}
\toprule
\textbf{Variant}
& \textbf{Unary}
& \textbf{Graph}
& \textbf{mIoU}$\uparrow$ \\
\midrule
Scribble only
& $\times$ & $\times$
& 33.2 \\
w/o unary
& $\checkmark$ & $\times$
& 79.6 \\
w/o graph
& $\times$ & $\checkmark$
& 61.0 \\
w/o CAC
& $\checkmark$ & $\checkmark$
& 80.3 \\
w/ uniform edges
& $\checkmark$ & $\checkmark$
& 80.3 \\
Full
& $\checkmark$ & $\checkmark$
& 85.3 \\
\bottomrule
\end{tabular}
\caption{
Component ablation under the NVOS benchmark. %
}
\label{tab:component_ablation}
\end{table}

As shown in Table~\ref{tab:component_ablation}, we conduct ablation studies under the NVOS benchmark to analyze the impact of different components of our method on segmentation quality.
We organize our method into unary evidence modeling and graph propagation.
Using scribbles alone yields only 33.2 mIoU, confirming that sparse scribbles are too incomplete to directly recover an object.
Adding graph propagation improves the result to 61.0 mIoU, showing that graph connectivity serves as an effective prior for propagating user scribbles, but remains insufficient without stronger evidence modeling approaches.
Combining scribble lifting with unary modeling reaches 79.6 mIoU, indicating that the lifted labels provide effective supervision for learning evidence likelihoods and the contrast between the background and the foreground objects.

\subsection{Qualitative Results}
\label{sec:qualitative_results}

\begin{figure}[t]
    \centering
    \includegraphics[width=1\linewidth]{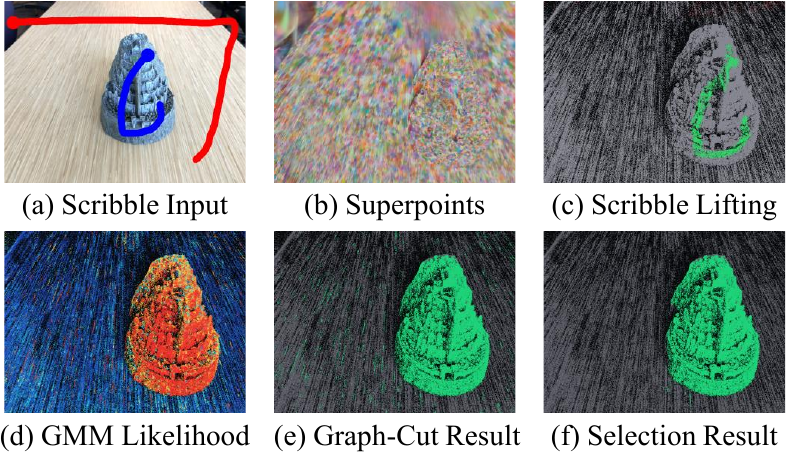}
    \caption{
    Visualization of the intermediate steps. %
    }
\label{fig:pipeline_intermediate}
\end{figure}

This section provides a qualitative analysis of GaussianSelector, showing that GaussianSelector enables effective object selection on the 3DGS superpoint graph from a small number of sparsely captured views.
We compare our method with GaussianCut~\cite{jain2024gaussiancut} under progressively enriched sparse-view input conditions.

\begin{figure}[h]
    \centering
    \includegraphics[width=\linewidth]{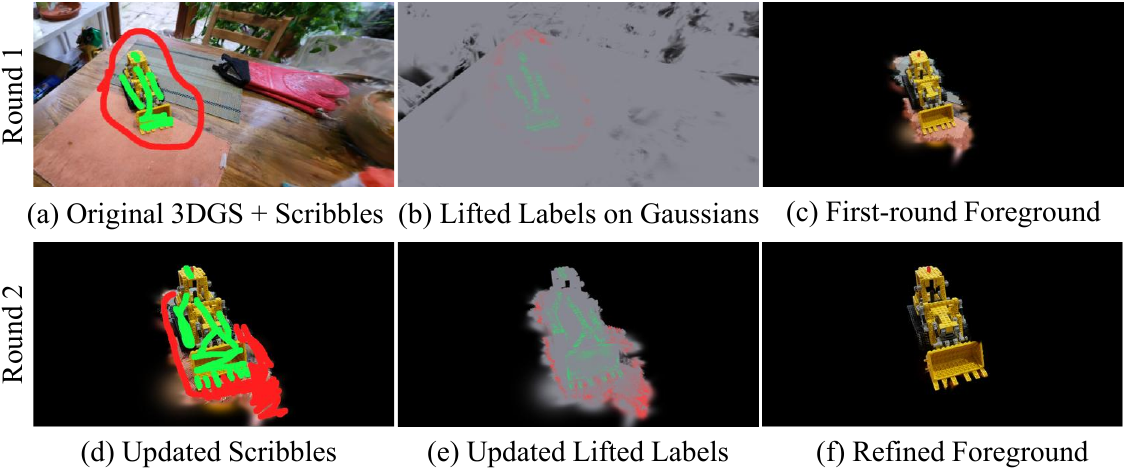}
    \caption{
    User interaction demonstration for human-in-the-loop object selection workflow.
    }
    \label{fig:ui_workflow}
\end{figure}

Fig.~\ref{fig:comp2} shows the object selection results under different sparse-view input settings. 
With limited NVOS scribbles, GaussianSelector already achieves competitive results, while additional human-in-the-loop refinement further improves boundary consistency and reduces artifacts. In contrast, GaussianCut is more sensitive to input quality, leading to noticeable degradation when fewer views or less informative inputs are provided. These results highlight the effectiveness of iterative refinement for fine-grained object selection. 
Fig.~\ref{fig:pipeline_intermediate} visualizes each step of GaussianSelector, showing how the user inputs progressively refine the selection.
Fig.~\ref{fig:ui_workflow} illustrates a human-in-the-loop interaction scenario of GaussianSelector.
As our method is characterized by fast inference on the 3DGS superpoint graph, we exploit this workflow to enable iterative refinement of object selection with fast and lightweight feedback, which is particularly useful for handling sophisticated structure and appearance patterns.

\subsection{User Study}
We conducted a user study with 12 participants on two object selection tasks. Participants rated intent matching, blind visual satisfaction, and waiting acceptability using 7-point Likert scales, and completed the standard System Usability Scale (SUS) questionnaire to assess overall usability.
As shown in Table~\ref{tab:user_study}, our scribble-based interaction provides more precise user control. Moreover, the substantially shorter response time enables users to iteratively refine their inputs based on immediate visual feedback, leading to markedly better waiting acceptability and overall usability.

\begin{table}[h]
\centering
\small
\begin{tabular}{@{}lccc@{}}
\toprule
 & \textbf{Ours} & \textbf{GaussianCut} & \textbf{FlashSplat} \\
\midrule
Intent matching & \textbf{6.3$\pm$0.6} & 5.8$\pm$0.8 & 5.8$\pm$0.7 \\
Visual satisfaction & 5.9$\pm$0.6 & \textbf{6.0$\pm$0.7} & 5.9$\pm$0.7 \\
Wait acceptability & \textbf{6.5$\pm$0.5} & 3.4$\pm$1.1 & 3.9$\pm$1.2 \\
SUS (0--100) & \textbf{84.2$\pm$6.8} & 63.5$\pm$9.4 & 60.8$\pm$10.2 \\
\bottomrule
\end{tabular}
\caption{
User study results (mean $\pm$ SD, best in bold).
}
\label{tab:user_study}
\end{table}
\section{Conclusion}

In this work, we present GaussianSelector, a plug-and-play and neural network-free framework that reformulates 3DGS interactive object selection as graph-based abstraction and evidence modeling directly in the native 3DGS space.
Experiments show that GaussianSelector achieves comparable performance with state-of-the-art 3DGS-based methods in the human-in-the-loop workflow while being significantly more efficient in computational cost and VRAM usage.
\clearpage

\bibliography{aaai2027}

\end{document}